%% file: main.tex
\documentclass{article}

\usepackage{iclr2027_conference,times}

\newcommand{\fitwidth}[1]{%
  \resizebox{\ifdim\width>\linewidth \linewidth\else\width\fi}{!}{#1}}
\usepackage{amsmath,amssymb}
\usepackage{array,booktabs,tabularx}
\usepackage{enumitem}
\usepackage{microtype}
\usepackage{xcolor}
\usepackage{tikz}
\usetikzlibrary{arrows.meta,calc,fit,positioning}
\usepackage{float}
\usepackage{algorithm}
\usepackage{algorithmic}
\usepackage[colorlinks=true,citecolor=blue!60!black,linkcolor=blue!60!black,urlcolor=blue!60!black]{hyperref}
\usepackage{graphicx}
\usepackage{natbib}
\usepackage{colortbl}
\makeatletter
\g@addto@macro\normalsize{%
  \setlength{\abovedisplayskip}{6pt plus 2pt minus 2pt}%
  \setlength{\belowdisplayskip}{6pt plus 2pt minus 2pt}%
  \setlength{\abovedisplayshortskip}{3pt plus 2pt}%
  \setlength{\belowdisplayshortskip}{4pt plus 2pt minus 2pt}%
}
\makeatother

\definecolor{pmblue}{HTML}{1971C2}
\definecolor{pmbluefill}{HTML}{E7F5FF}
\definecolor{pmgreen}{HTML}{2B8A3E}
\definecolor{pmgreenfill}{HTML}{EBFBEE}
\definecolor{pmorange}{HTML}{E67700}
\definecolor{pmorangefill}{HTML}{FFF4E6}
\definecolor{pmpurple}{HTML}{7048E8}
\definecolor{pmpurplefill}{HTML}{F3F0FF}
\definecolor{pmred}{HTML}{C92A2A}
\definecolor{pmredfill}{HTML}{FFF5F5}
\definecolor{pmteal}{HTML}{0B7285}
\definecolor{pmtealfill}{HTML}{E3FAFC}
\definecolor{pmgray}{HTML}{495057}
\definecolor{pmgrayfill}{HTML}{F1F3F5}
\tikzset{
  pm box/.style={draw=pmgray, rounded corners=3pt, line width=0.75pt,
    align=center, inner sep=4pt, font=\sffamily\scriptsize},
  pm small/.style={pm box, font=\sffamily\tiny, inner sep=3pt},
  pm arrow/.style={-{Latex[length=2.0mm,width=1.35mm]}, draw=pmgray,
    line width=0.75pt},
  pm pending/.style={dash pattern=on 3pt off 2pt},
  pm label/.style={font=\sffamily\bfseries\scriptsize, text=pmgray},
  pm note/.style={font=\sffamily\tiny, align=center, text=pmgray}
}

\newcommand{\payoffpanel}[1]{%
  \begin{tikzpicture}[x=0.0470cm, y=0.0680cm]
    \foreach \gx in {80,100,120} \draw[gray!15,very thin] (\gx,-10) -- (\gx,20);
    \foreach \gy in {-10,0,10,20} \draw[gray!15,very thin] (75,\gy) -- (125,\gy);
    \draw[gray,dashed,thin] (75,0) -- (125,0);
    \draw[gray,thin] (75,-10) rectangle (125,20);
    \foreach \gx in {80,100,120} {%
      \draw[gray,thin] (\gx,-10) -- (\gx,-8.2);
      \node[font=\tiny,below=0.5pt] at (\gx,-10) {\gx};}
    \foreach \gy in {-10,0,10,20} {%
      \draw[gray,thin] (75,\gy) -- (77.2,\gy);
      \node[font=\tiny,left=0.5pt] at (75,\gy) {\gy};}
    \draw[blue!70!black,very thick,line join=round,line cap=round] #1;
    \node[font=\tiny,below=6.5pt] at (100,-10) {$S_T$};
  \end{tikzpicture}}
\newcommand{\payofftitle}[1]{%
  \parbox[b][30pt][b]{3.0cm}{\centering\scriptsize #1\par}}

\newcommand{\system}{\textsc{Sota}}
\newcommand{\sotarow}{\rowcolor{gray!12}}

\newcommand{\code}[1]{\texttt{#1}}

\newcolumntype{Y}{>{\raggedright\arraybackslash}X}

\newcommand{\blfootnote}[1]{%
  \begingroup
  \renewcommand{\thefootnote}{}\footnote{#1}%
  \addtocounter{footnote}{-1}%
  \endgroup}

\title{\system: Stock Options Trading Agents Guided by\\
Option-Implied Return Distributions}
\author{Yizhen Xie$^{*}$ \\
Carnegie Mellon University \\
\And
Mengyang Liu$^{\dagger}$ \\
Amazon}
\date{September 21, 2026}

\iclrfinalcopy

\begin{document}
\maketitle

% The style's final branch sets \lhead{Published as a conference paper at ICLR
% 2027}, which this paper is not. Two details make this override fiddly:
%
% It must come *after* \maketitle, not via \AtBeginDocument -- the style assigns
% \lhead inside \@maketitle, so anything set earlier is overwritten when the
% title is typeset.
%
% And it must be guarded. Unguarded it also replaces the *blind* branch's
% "Under review as a conference paper at ICLR 2027", which is the header an ICLR
% submission is supposed to carry; removing \iclrfinalcopy would then hide the
% names but leave a non-standard header. Verified by building both ways.
\ificlrfinal\lhead{}\fi

% Affiliation disclosure.  Guarded by \ificlrfinal so it is suppressed
% automatically while the paper is anonymous and appears only once
% \iclrfinalcopy is set -- naming an employer in a double-blind submission
% would break anonymity, so the guard is the safety, not a convenience.
\ificlrfinal
% The affiliation itself is now stated in the author block, so this note carries
% only the independence disclaimer.  The "errors are our own" clause moved to
% the Acknowledgments, which is its conventional home and where the author's own
% wording already places it -- keeping both would print it twice.
\blfootnote{$^{*}$ Correspondence to: yizhenxi@andrew.cmu.edu.}
\blfootnote{$^{\dagger}$ Work conducted outside the author's role at Amazon.}
\fi

\begin{abstract}
As option markets grow and AI advances, agentic systems for option trading are gaining increasing attention. Language-model-based agents can reason over contextual information such as news, but option trading presents a particularly challenging decision problem: a single stock can have thousands of contracts, and the agent must decide both which contracts to trade and how to combine them. Existing approaches often sidestep this complexity by restricting the policy to a fixed strategy structure, such as a straddle, limiting their ability to switch strategies as market conditions change. We present SOTA (Stock Options Trading Agents), an agentic trading framework for structured option-strategy selection. SOTA abstracts the large option universe into strategy-level decisions while deterministic resolvers handle portfolio implementation. We develop SOTA by post-training Qwen3.8-27B with supervised fine-tuning followed by reinforcement learning. SOTA is evaluated on options on nine large-cap U.S. equities and SPY against rule-based and machine-learning strategy selectors in the same trading environment. Over a six-month out-of-sample period, SOTA earns an 18.3\% total return with a Sharpe ratio of 1.60 and a maximum drawdown of 8.96\%. We also document an asymmetric role of news: news improves frontier-teacher trajectories, but retaining news during reinforcement learning reduces out-of-sample return from 18.3\% to $-$2.7\%.
\end{abstract}

\input{introduction}

\input{background}

\input{method}

\input{experiments}

\section{Ablation Studies}

We examine two components of the training pipeline: the role of news in the
policy information set and the contribution of reinforcement learning after
supervised fine-tuning. Table~\ref{tab:ablation} summarizes the results.

\subsection{The Role of News in the Information Set}

To reduce temporal and identity leakage, we anonymize the information presented
to the frontier teacher by masking stock identity, absolute price levels, and
calendar time, while retaining market states and contemporaneously available
news. We find that including anonymized news improves the quality of the
frontier-model trajectories used for supervised fine-tuning. After filtering
trajectories by annualized Sharpe ratio, Appendix~\ref{app:family-breakdown}
reports performance across the retained trajectories by strategy family.

We then ask whether news remains useful during reinforcement learning. Holding
the supervised checkpoint fixed, we compare reinforcement learning with and
without contemporaneous news in the policy information set. The policy trained
without news achieves an 18.32\% total return and a Sharpe ratio of 1.60, with
a maximum drawdown of 8.96\%. Retaining news during reinforcement learning
lowers total return to $-2.72\%$ and raises maximum drawdown to 31.69\%.

These results point to an asymmetric role of textual information: news is useful
for constructing frontier-model supervision, but retaining it during policy
optimization does not improve out-of-sample performance.

\subsection{The Role of Reinforcement Learning}

We next evaluate the supervised policy before reinforcement learning. Because
the SFT-only policy and the RL-with-news policy use the same information set,
their comparison isolates the contribution of reinforcement learning.

Reinforcement learning raises total return from $-10.19\%$ to $-2.72\%$ and
improves the Sharpe ratio from $-1.25$ to $-0.16$, although volatility and
maximum drawdown also increase. This indicates that reinforcement learning
improves the supervised policy under the same information set, while the
performance of the full method additionally depends on removing news during
policy optimization.

\input{ablation}

\section{Conclusion}

We introduce SOTA, a language-model post-training framework for structured option-strategy selection. By separating strategy-level reasoning from contract-level implementation, SOTA allows the learned policy to choose among heterogeneous option strategies while deterministic resolvers handle the implementation. Across 10 actively traded option underlyings, the post-trained policy improves out-of-sample portfolio performance relative to the non-LLM strategy-selection baselines and the fine-tuned language policy. We also document an asymmetric role of textual information: news improves frontier-model supervision, but retaining news during reinforcement learning does not improve out-of-sample performance. More broadly, the results suggest that information useful for constructing expert supervision need not remain equally useful during downstream policy optimization. Future work can extend the learned decision space to contract selection, position sizing, and dynamic hedging, as well as explore alternative representations of structured market and textual information.

% Acknowledgments.  Guarded by \ificlrfinal for the same reason as the
% affiliation footnote on page 1: naming a supercomputing center identifies the
% allocation and therefore the institution, which a double-blind submission must
% not do.  Suppressed while the paper is anonymous; appears once \iclrfinalcopy
% is set in the preamble.
\ificlrfinal
\section*{Acknowledgments}

We gratefully acknowledge the Pittsburgh Supercomputing Center for providing
computational resources that supported this research. The options and market
data used in this study were obtained under licenses from commercial data
providers. All errors and omissions remain the responsibility of the authors.
\fi

\bibliographystyle{iclr2027_conference}
\bibliography{references}

\appendix
\input{appendix}

\end{document}

%% file: introduction.tex
\section{Introduction}

Option trading has expanded rapidly in recent years \citep{bryzgalova2023retail}, creating a natural setting for increasingly capable trading agents. Recent language-model trading systems formulate financial decision-making as sequential interaction with a market environment, where an agent observes market information and portfolio state and chooses actions such as buy, sell, or hold \citep{yang2023fingpt,yu2025finmem,zhang2024finagent,xiao2024tradingagents}. Extending this formulation to options is difficult because the decision space is much larger and changes over time.

For a single underlying, an agent may face thousands of contracts that differ in strike, maturity, and option type. Existing approaches reduce this complexity either by predicting returns for individual contracts \citep{bali2023option} or by restricting the trading problem to a single strategy class, such as volatility trading \citep{sheu2011effective,chen2025ophr}. These formulations are useful, but they do not directly support dynamic selection across heterogeneous payoff structures.

Reinforcement learning (RL) is well suited to sequential trading problems with transaction costs, discrete rebalancing, and portfolio-state dependence \citep{sutton2018reinforcement,mnih2015human,buehler2019deephedging,chen2025ophr}. It has been applied to portfolio management, stock trading, and high-frequency trading \citep{jiang2017cryptocurrency,yang2020ensemble,briola2021deep,qin2024earnhft,zong2024macrohft}. In options, however, learning-based methods have focused primarily on hedging \citep{buehler2019deephedging} or on learning positions within a fixed volatility structure such as an at-the-money straddle \citep{tan2024deeplearningoptions,chen2025ophr}. Selecting among heterogeneous option strategies remains challenging.

In this paper, we present \textbf{SOTA (Stock Options Trading Agents)}, an agentic trading framework for structured option-strategy selection. SOTA separates strategy-level reasoning from contract-level implementation: the language model selects among nine economically meaningful strategy families and their parameters, while deterministic resolvers map each decision into exact contracts, position sizes, and hedges. This structured strategy-selection interface reduces the decision space without restricting the agent to a single payoff structure. The policy can also combine structured market information with contextual signals such as news before producing executable option positions.

The primary contributions of our work are:
\begin{itemize}
    \item We formulate option trading as structured strategy selection rather than contract-level choice. The agent selects among nine strategy families and their parameters, while deterministic resolvers map these decisions into contracts, position sizes, and hedges. This reduces a large contract-level decision problem to a compact strategy space spanning directional, volatility, skew, and curvature trades.

    \item We construct a point-in-time trading environment whose state summarizes the option-implied return distribution, and train Qwen3.8-27B using supervised fine-tuning on anonymized frontier-model trajectories followed by reinforcement learning on portfolio returns net of transaction costs.

\item SOTA earns an 18.3\% total return with a Sharpe ratio of 1.60 over a six-month out-of-sample period on options on nine large-cap U.S.\ stocks and SPY, while all rule-based and machine-learning baselines generate negative returns. Ablations show that reinforcement learning improves on supervised fine-tuning alone, while news helps the frontier teacher but degrades performance when retained during reinforcement learning.
\end{itemize}

%% file: background.tex
\section{Related Work}

\subsection{Agentic Trading}
\label{sec:agentic-trading}

Agentic trading systems treat trading as repeated interaction with a market: at each step the agent observes prices and its portfolio, takes an action, and receive feedback through profit and loss. Early systems learn this policy with reinforcement learning. They have been applied to trading stocks \citep{yang2020ensemble}, cryptocurrency \citep{jiang2017cryptocurrency,qin2024earnhft,zong2024macrohft}, and high frequency trading using limit-order-book data \citep{briola2021deep}, and shared environments such as FinRL-Meta make these settings easy to benchmark \citep{liu2022finrlmeta}. These approaches operate primarily on numerical market data.

Language-model agents add the ability to process news, filings, and other text. Financial language models adapt general-purpose models to financial text \citep{yang2023fingpt,yang2023investlm}. Trading agents built on such models add memory of past events \citep{yu2025finmem}, multimodal inputs and tool use \citep{zhang2024finagent}, or multi-agent roles such as analyst, researcher, trader, and risk manager \citep{yu2024fincon,xiao2024tradingagents}, and agent platforms and market simulators support these designs \citep{yang2024finrobot,zhang2024stockagent}. When trading on their own, the performance of language-model agents remains mixed: on real stock data, StockBench finds that most language-model agents struggle to outperform a simple buy-and-hold baseline \citep{chen2025stockbench}.

In both lines of work, the action space is small and fixed: buy, sell, or hold a single stock or cryptocurrency, or choose portfolio weights over a fixed set of assets. Options do not fit this mold. Each underlying may have thousands of contracts, the available set changes as contracts expire and new ones are listed, and a position often combines several contracts. SOTA keeps the agentic loop but changes the action: the agent selects an option strategy and its parameters, and deterministic tools turn that decision into contracts, position sizes, and hedges.

\subsection{Option-Implied Return Distributions}
\label{sec:options-bsm}

Option prices encode the market's distribution of future returns \citep{breeden1978prices}. Implied volatilities across strikes and maturities reveal the volatility, skewness, and kurtosis of this distribution \citep{corrado1996skewness,dennis2002riskneutral,bakshi2003stock,dumas1998implied,cont2002dynamics,zhang2008smirk}. These option-implied measures are informative about future outcomes. Implied volatility predicts realized volatility \citep{christensen1998relation}, option trading volume carries information about future stock prices and volatility \citep{easley1998optionvolume,pan2006information,ni2008volatility}, and volatility-related characteristics predict the cross-section of option returns \citep{goyal2009crosssection,cao2013crosssection}, a relation that machine-learning models exploit at scale \citep{bali2023option}. Our market state is built from these measures.

\subsection{Option Trading Strategies}

Option strategies combine contracts to create different payoff exposures.
A second-order approximation of option value gives
\[
dV \approx \Delta\, dS
+ \frac{1}{2}\Gamma(dS)^2
+ \nu\, d\sigma
+ \Theta\, dt.
\]
Here, delta measures directional exposure, gamma reflects convexity with respect to the underlying price, theta represents time decay, and vega measures sensitivity to implied volatility. Appendix~\ref{app:greeks} gives their Black--Scholes--Merton formulas \citep{black1973pricing,merton1973rational}.
These sensitivities motivate four broad classes of trading strategies in our
framework: directional, volatility, skewness, and curvature strategies.

\paragraph{Directional strategies.}
Directional strategies express a view on whether the underlying asset will rise
or fall. Their payoffs are driven mainly by net delta exposure, while different
option structures allow the same directional view to be expressed with different
degrees of leverage, asymmetry, and downside protection.

\paragraph{Volatility strategies.}
Volatility strategies express a view on how much the underlying asset will move,
rather than on the direction of the move. Long-volatility positions benefit from
large price movements, while short-volatility positions benefit when price
movements remain limited. Their performance depends on the trade-off between
convexity, time decay, and changes in option prices
\citep{bakshi2003deltahedged,carr2009variance,goyal2009crosssection}.

\paragraph{Skewness strategies.}
Skewness strategies express a view on whether downside and upside risks are priced
differently. They compare the relative value of put and call options away from the
current stock price and can profit when this asymmetry changes
\citep{bakshi2003stock,garleanu2009demand,kozhan2013skew}.

\paragraph{Curvature strategies.}
Curvature strategies express a view on how option prices differ between contracts
near the current stock price and contracts further away from it. They profit from
changes in the relative pricing of central and more extreme strike prices
\citep{breeden1978prices,zhang2008smirk}.

SOTA performs structured strategy selection across nine strategy families that
span these four categories. Appendix~\ref{app:payoffs} illustrates representative payoff
structures for each family.

%% file: method.tex
\section{SOTA: Structured Strategy Selection for Stock Option Trading Agents}
\label{sec:method}

Option trading presents a large and dynamically changing decision problem:
for a single underlying, an agent may face thousands of contracts that differ
in strike, maturity, and option type. Direct contract-level selection is therefore
difficult in a time-sensitive trading setting.

\begin{figure}[H]
    \centering
    % Source: figs/cartoon_v1b.svg (build the PDF with rsvg-convert, see its header).
    % The previous figure was sota.png, with the caption kept below.
    \includegraphics[width=\linewidth]{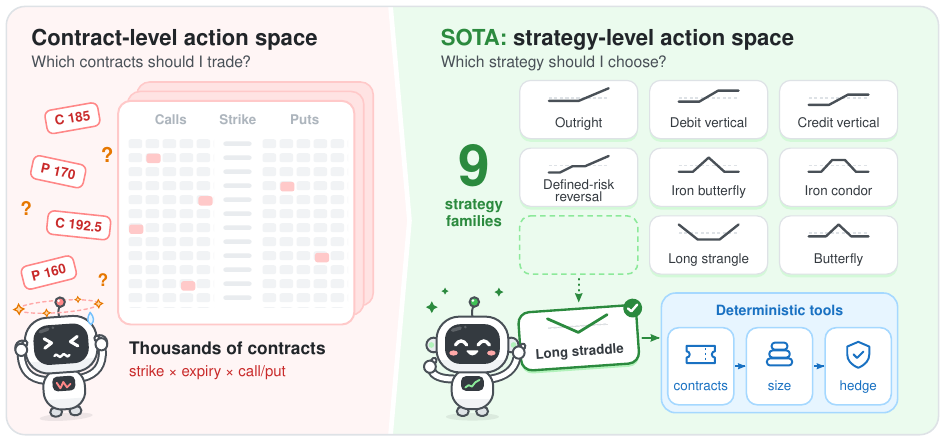}
    \caption{From contract-level to strategy-level decisions. Left: at the contract level, an agent must choose among thousands of listed contracts that differ in strike, expiry, and call/put type. Right: SOTA instead has the agent select one of nine option-strategy families and its parameters (here, a long straddle), and deterministic tools resolve that choice into exact contracts, a position size, and hedge trades.}
    % \caption{Overview of SOTA. A frontier LLM uses structured market states and contemporaneous news to generate supervised trading trajectories. The smaller language model is initialized by supervised fine-tuning and then optimized by reinforcement learning in the portfolio environment. The policy makes strategy-level decisions, and deterministic resolvers map them into executable portfolio positions.}
    \label{fig:sota}
\end{figure}

SOTA formulates this problem as structured strategy selection. The language model
chooses an option strategy and its parameters, while deterministic resolvers handle
contract selection, position sizing, hedging, and portfolio implementation. This
separation reduces the contract-level decision problem while preserving flexibility
across heterogeneous payoff structures. Figure~\ref{fig:sota} summarizes the
framework.

\input{problem}

We train the policy in two stages. Supervised fine-tuning on frontier-model trading trajectories provides an initial strategy-selection policy, after which GRPO \citep{shao2024deepseekmath} optimizes the policy for portfolio performance. Appendix~\ref{app:training-algorithm} describes the full procedure, and Appendix~\ref{app:training-hparams} reports the hyperparameters.

%% file: problem.tex
\subsection{Problem Formulation}
\label{sec:problem}

\textbf{Trading environment.}
The agent trades options on a set of underlyings $\mathcal{U}$ over decision dates $t=0,1,\dots,T$. On date $t$, underlying $u\in\mathcal{U}$ has price $S_t^u$ and a listed option chain $\mathcal{C}_t^u$. Each contract $c\in\mathcal{C}_t^u$ is defined by its strike, expiration date, and type (call or put), and has price $p_t^c$ per contract, taken at the bid--ask midpoint. A chain holds thousands of contracts with different nonlinear exposures, and it changes over time as contracts expire and new ones are listed.

The portfolio entering date $t$ holds $q_t^c$ contracts of each option $c$, $h_t^u$ shares of each underlying for delta hedging, and cash $B_t$. Its marked-to-market value is
\begin{equation}
V_t
=
\underbrace{\sum_{u\in\mathcal{U}}\sum_{c\in\mathcal{C}_t^u} q_t^c\,p_t^c}_{V_t^{\mathrm{options}}}
+
\underbrace{\sum_{u\in\mathcal{U}} h_t^u S_t^u}_{V_t^{\mathrm{underlying}}}
+
\underbrace{B_t}_{V_t^{\mathrm{cash}}}.
\label{eq:portfolio-value}
\end{equation}
Trading is sequential. At each date $t$, the agent observes the market and its portfolio and issues a trading decision $a_t$. The environment converts $a_t$ into trades $\Delta q_t^c$ and $\Delta h_t^u$, executes them at midpoint prices, and charges fees and commissions $\mathrm{TC}_t$. Holdings and cash after trading are
\begin{equation}
q_t^{c,+}=q_t^c+\Delta q_t^c,
\qquad
h_t^{u,+}=h_t^u+\Delta h_t^u,
\qquad
B_t^{+}
=
B_t
-\sum_{u\in\mathcal{U}}\Bigl(\sum_{c\in\mathcal{C}_t^u}\Delta q_t^c\,p_t^c+\Delta h_t^u S_t^u\Bigr)
-\mathrm{TC}_t .
\label{eq:portfolio-update}
\end{equation}
These holdings carry over to date $t+1$, where they are marked at the new prices and expiring contracts are settled. The change from $V_t$ to $V_{t+1}$ therefore reflects option price moves, hedge positions, transaction costs, and settlements.

Choosing $\Delta q_t^c$ directly over every listed contract creates a large, dynamically changing decision space. We instead represent each decision $a_t$ as a standardized option combination from an economically meaningful strategy family, and deterministic resolvers translate it into contract-level trades.

\textbf{Markov decision process.}
We formulate option-strategy selection as a Markov decision process,
\begin{equation}
\mathcal{M}
=
\langle
\mathcal{S},
\mathcal{A},
\mathcal{T},
\mathcal{R},
\gamma
\rangle,
\label{eq:mdp}
\end{equation}
where $\mathcal{S}$ is the state space, $\mathcal{A}$ the structured strategy space, $\mathcal{T}$ the transition mechanism, $\mathcal{R}$ the reward function, and $\gamma$ the discount factor.

\textbf{State.}
At each trading date $t$, the agent observes a point-in-time state
\begin{equation}
s_t
=
\left(
\mathbf{M}_t,
\mathbf{N}_t,
\mathbf{P}_t,
\mathbf{A}_t
\right),
\label{eq:state}
\end{equation}
where $\mathbf{M}_t$ contains market features, $\mathbf{N}_t$ point-in-time textual summaries of news and filings, $\mathbf{P}_t$ open option positions, and $\mathbf{A}_t$ portfolio-level information. Portfolio state allows each decision to condition on exposures and outcomes inherited from earlier trades.

We construct the market state from variables identified by the option-pricing and option-return literature as relevant to variation in option returns \citep{goyal2009crosssection,cao2013crosssection,bakshi2003stock,pan2006information}. For each underlying, the state contains 10 features: underlying return, realized volatility, implied-volatility level and change, the implied-minus-realized volatility wedge, term structure, skewness, curvature, option flow, and changes in open interest. Appendix~\ref{app:market-features} gives precise definitions.

\textbf{Structured strategy selection.}
Instead of choosing directly among listed contracts, the policy selects structured trading strategies. At each decision time, the policy outputs a set of actions: open a new position ($a_t^{\mathrm{open}}$), close an existing position ($a_t^{\mathrm{close}}$), roll an existing position to a later expiry ($a_t^{\mathrm{roll}}$), or leave the portfolio unchanged ($a_t^{\mathrm{hold}}$).

An action $a_t$ is specified as
$\left(u,f,\omega,\theta,\kappa\right)$,
where $u$ is the underlying, $f$ the strategy family, $\omega$ the long/short orientation, $\theta$ the tenor bucket, and $\kappa$ the strategy's coordinates on the delta grid.

To avoid querying prices for the full option chain, the agent first proposes up to six candidate strategies for each underlying. We resolve and price only these candidate packages before the final trading decision. This two-stage mechanism preserves price-aware strategy selection without placing thousands of contracts in the model context.

The learned policy therefore focuses on strategy selection, while deterministic resolvers handle contract selection, portfolio sizing, and delta hedging. Appendix~\ref{app:formulas} reports the rules implemented by the latter.

\textbf{Transition.}
The transition kernel $\mathcal{T}(s_{t+1}\mid s_t,a_t)$ captures changes in market state, news and filings, existing positions and their risk exposures, and portfolio value net of transaction costs, with holdings and cash evolving as in Equation~\ref{eq:portfolio-update}.

\textbf{Reward.}
The one-step reward is the change in log portfolio value,
\begin{equation}
r_t
=
\log V_{t+1}
-
\log V_t.
\label{eq:reward}
\end{equation}

%% file: experiments.tex
\section{Experiments}
\label{sec:experiments}

\subsection{Experiment Setup}
\label{sec:setup}

\paragraph{Datasets.}
We evaluate SOTA on 10 actively traded underlyings: SPY and nine large-capitalization U.S.\ equities (AAPL, AMZN, GOOGL, META, MSFT, MU, NVDA, PLTR, and TSLA). The option data include OPRA trades and quotes and OptionMetrics. Underlying stock prices are CRSP daily open and close prices for the nine equities and SpiderRock underlying marks for SPY. News comes from Massive.com's news API, and regulatory filings come from SEC EDGAR. Table~\ref{tab:splits} reports the chronological sample splits.

\begin{table}[H]
\centering
\small
\setlength{\tabcolsep}{3pt}
\caption{Chronological data splits.}
\begin{tabularx}{\linewidth}{@{}llY@{}}
\toprule
Phase & Dates & Role \\
\midrule
Supervised fine-tuning
& 2024-09-03 to 2024-11-29
& Frontier-model generation and supervised fine-tuning \\

Reinforcement learning
& 2024-12-02 to 2025-02-28
& Reinforcement learning to optimize the policy \\

Test
& 2025-03-03 to 2025-08-29
& Out-of-sample evaluation of the policy and baselines \\
\bottomrule
\end{tabularx}
\label{tab:splits}
\end{table}

\paragraph{Evaluation metrics.}
We evaluate portfolio performance using eight financial metrics: total return (TR), annualized Sharpe ratio (ASR), annualized Calmar ratio (ACR), annualized Sortino ratio (ASoR), annualized volatility (AVOL), maximum drawdown (MDD), win rate (WR), and profit/loss ratio (PLR). TR measures cumulative return over the evaluation period. AVOL is the annualized standard deviation of daily returns, and MDD is the largest peak-to-trough decline in portfolio value. ASR measures return relative to total volatility, ACR relative to maximum drawdown, and ASoR relative to downside deviation. WR is the fraction of completed trades with positive profit, and PLR is the average profit of winning trades relative to the average loss of losing trades. Appendix~\ref{app:evaluation-metrics} gives detailed definitions.

\paragraph{Baselines.}
We compare SOTA with conventional rule-based and supervised machine-learning policies for option-strategy selection. The rule-based baselines include equal-weighted strategy allocation, a GARCH-based rule \citep{bollerslev1986generalized}, and a threshold policy based on the structured market features. The GARCH rule combines a volatility forecast with contemporaneous option-market information. The supervised machine-learning baselines predict returns from the numerical market state and map those predictions into strategy choices using predetermined implementation rules. All feasible policies use the same test dates, execution model, transaction-cost assumptions, portfolio constraints, and deterministic resolvers, so performance differences reflect the strategy-selection rule rather than portfolio implementation. Appendix \ref{app:hyperparameters} provides the specifications of the baselines.

\input{baselines}

\subsection{Baseline Comparison}
\label{sec:main-results}

Table~\ref{tab:baselines} compares SOTA with the rule-based and machine-learning baselines over the test period. SOTA earns a total return of 18.32\%, an annualized Sharpe ratio of 1.60, and a maximum drawdown of 8.96\%. All rule-based and supervised machine-learning baselines produce negative total and risk-adjusted returns in the same evaluation. The equal-weighted rule has relatively low volatility but negative return, while the GARCH and threshold rules exhibit larger losses and drawdowns. The GBDT and logistic policies lose 49.78\% and 55.45\%, respectively. In this evaluation, SOTA is the only feasible policy with positive total and risk-adjusted returns, indicating that the post-trained language policy captures strategy-selection patterns not reproduced by the fixed rules or supervised mappings considered here.

The final row of Table~\ref{tab:baselines} reports a hindsight oracle. At each decision time, it selects the best-performing strategy from the same candidate set using realized future outcomes. The oracle is therefore not tradable, but it uses the same transaction costs, sizing rule, position limits, and hedging rule as the feasible policies. Its return provides a reference for the profitability available within the structured strategy space.

%% file: baselines.tex
\begin{table}[!t]
\centering
\small
\setlength{\tabcolsep}{2.5pt}
\renewcommand{\arraystretch}{0.94}

\caption{Baseline comparison for option-strategy selection policies}
\label{tab:baselines}

\resizebox{\linewidth}{!}{%
\begin{tabular}{@{}lcccccccc@{}}
\toprule
& Profit & \multicolumn{3}{c}{Risk-adjusted profit}
& \multicolumn{2}{c}{Risk}
& \multicolumn{2}{c}{Trade} \\
\cmidrule(lr){2-2}
\cmidrule(lr){3-5}
\cmidrule(lr){6-7}
\cmidrule(lr){8-9}
Policy
& TR(\%)$\uparrow$
& ASR$\uparrow$
& ACR$\uparrow$
& ASoR$\uparrow$
& AVOL(\%)$\downarrow$
& MDD(\%)$\downarrow$
& WR(\%)$\uparrow$
& PLR$\uparrow$ \\
\midrule

\multicolumn{9}{@{}l}{\emph{Rule-based policies}} \\
\quad Equal-weighted rule
& $-5.22$ & $-13.46$ & $-1.98$ & $-11.54$
& $\mathbf{0.81}$ & $\mathbf{5.22}$ & $31.11$ & $1.08$ \\

\quad GARCH rule
& $-49.24$ & $-6.27$ & $-1.52$ & $-6.25$
& $22.17$ & $49.42$ & $29.85$ & $0.75$ \\

\quad Threshold rule
& $-28.46$ & $-1.95$ & $-1.50$ & $-3.50$
& $35.20$ & $33.16$ & $\mathbf{41.31}$ & $1.89$ \\

\addlinespace[4pt]

\multicolumn{9}{@{}l}{\emph{Machine-learning policy}} \\
\quad GBDT
& $-49.78$ & $-8.55$ & $-1.52$ & $-8.18$
& $16.51$ & $49.78$ & $24.07$ & $1.23$ \\

\quad Logistic
& $-55.45$ & $-11.63$ & $-1.46$ & $-9.70$
& $14.24$ & $55.47$ & $20.42$ & $1.00$ \\

\addlinespace[4pt]

\multicolumn{9}{@{}l}{\emph{Language policy}} \\
\sotarow\quad SOTA
& $\mathbf{18.32}$ & $\mathbf{1.60}$ & $\mathbf{4.59}$ & $\mathbf{4.77}$
& $21.52$ & $8.96$ & $29.17$ & $\mathbf{4.68}$ \\

\addlinespace[4pt]

\multicolumn{9}{@{}l}{\emph{Hindsight reference}} \\
\quad Oracle
& $608.31$ & $6.28$ & $740.22$ & $35.87$
& $63.91$ & $7.32$ & -- & -- \\

\bottomrule
\end{tabular}%
}

\vspace{2pt}
\begin{minipage}{\linewidth}
\footnotesize
\textit{Notes:} Upward arrows indicate that higher values are better; downward arrows indicate that lower values are better. Bold marks the best value in each column among the tradable policies; the hindsight oracle is excluded. Appendix~\ref{app:evaluation-metrics} gives detailed definitions of the metrics.
\end{minipage}

\end{table}

%% file: ablation.tex
\begin{table}[H]
\centering
\small
\setlength{\tabcolsep}{2.5pt}
\renewcommand{\arraystretch}{0.94}

\caption{Ablation studies over the test period. Bold marks the best value in each column.}
\label{tab:ablation}

\fitwidth{%
\begin{tabular}{@{}lcccccccccc@{}}
\toprule
& \multicolumn{2}{c}{Factor}
& Profit & \multicolumn{3}{c}{Risk-adjusted profit}
& \multicolumn{2}{c}{Risk}
& \multicolumn{2}{c}{Trade} \\
\cmidrule(lr){2-3}
\cmidrule(lr){4-4}
\cmidrule(lr){5-7}
\cmidrule(lr){8-9}
\cmidrule(lr){10-11}
Variant
& RL & News
& TR(\%)$\uparrow$
& ASR$\uparrow$
& ACR$\uparrow$
& ASoR$\uparrow$
& AVOL(\%)$\downarrow$
& MDD(\%)$\downarrow$
& WR(\%)$\uparrow$
& PLR$\uparrow$ \\
\midrule

\multicolumn{11}{@{}l}{\emph{Full method}} \\
\sotarow\quad SOTA & \checkmark & $\times$
& $\mathbf{18.32}$ & $\mathbf{1.60}$ & $\mathbf{4.59}$ & $\mathbf{4.77}$
& $21.52$ & $\mathbf{8.96}$ & $29.17$ & $\mathbf{4.68}$ \\

\addlinespace[4pt]

\multicolumn{11}{@{}l}{\emph{Ablations}} \\
\quad w/ news during RL & \checkmark & \checkmark
& $-2.72$ & $-0.16$ & $-0.17$ & $-0.29$
& $36.00$ & $31.69$ & $\mathbf{38.89}$ & $1.36$ \\

\quad w/o RL (SFT only) & $\times$ & \checkmark
& $-10.19$ & $-1.25$ & $-1.27$ & $-1.91$
& $\mathbf{17.59}$ & $15.56$ & $37.50$ & $0.97$ \\

\bottomrule
\end{tabular}}

\end{table}

%% file: appendix.tex
 \section{Option Pricing and Greeks}
\label{app:greeks}

The Black-Scholes-Merton (BSM) Model is a seminal framework for pricing options.  Write $S$ for the underlying price, $K$ the strike,
$T$ the time to expiry in years, $r$ the risk-free rate, $q$ the continuous
dividend yield, and $\sigma$ the implied volatility.  

\subsection{Black--Scholes-Merton Pricing Formula}

With
\begin{equation}
d_1 = \frac{\ln(S/K) + \bigl(r - q + \tfrac12\sigma^{2}\bigr)T}{\sigma\sqrt{T}},
\qquad
d_2 = d_1 - \sigma\sqrt{T},
\label{eq:d1d2}
\end{equation}
the European call and put values are
\begin{align}
C &= S e^{-qT} \Phi(d_1) \;-\; K e^{-rT} \Phi(d_2), \label{eq:bs-call}\\
P &= K e^{-rT} \Phi(-d_2) \;-\; S e^{-qT} \Phi(-d_1). \label{eq:bs-put}
\end{align}
where $\Phi$ is the
standard normal distribution function. Implied volatility is defined as the volatility $\sigma$ that equates an option's market price with the theoretical value.  

\subsection{Option Greeks}

\begin{align}
\Delta &= \frac{\partial V}{\partial S}
  = e^{-qT} N(d_1) \ \ \text{(call)}, \qquad -e^{-qT} N(-d_1) \ \ \text{(put)},
  \label{eq:delta}\\[1mm]
\Gamma &= \frac{\partial^{2} v}{\partial S^{2}}
  = \frac{e^{-qT}\,\phi(d_1)}{S\,\sigma\sqrt{T}}, \label{eq:gamma}\\[1mm]
\nu &= \frac{\partial V}{\partial \sigma}
  = S e^{-qT}\,\phi(d_1)\,\sqrt{T}, \label{eq:vega}\\[1mm]
\Theta &= \frac{\partial V}{\partial t}
  = -\frac{S e^{-qT}\phi(d_1)\sigma}{2\sqrt{T}}
    - rKe^{-rT}N(d_2) + qSe^{-qT}N(d_1) \ \ \text{(call)}. \label{eq:theta}
\end{align}
where $\phi$ is the standard normal density.

\section{Strategy payoffs illustrated}
\label{app:payoffs}

This appendix illustrates the net payoff structure of each of the nine strategy families considered.
\begin{table}[tbp]
\centering
\caption{Expiration profit and loss of the nine option-strategy families, net of
the initial cost of the package\textsuperscript{\,a}}
\label{tab:option-payoff-profiles}
\vspace{0.6em}

\begin{tabular}{@{}c@{\hspace{0.18cm}}c@{\hspace{0.18cm}}c@{}}

\payofftitle{\textbf{[ol]} Outright\\
$+C_{100}$; debit $5$} &
\payofftitle{\textbf{[dv]} Debit vertical\\
$+C_{95}-C_{105}$; debit $5$} &
\payofftitle{\textbf{[cv]} Credit vertical\\
$-P_{105}+P_{95}$; credit $5$} \\[0.2em]

\payoffpanel{(75,-5) -- (100,-5) -- (125,20)} &
\payoffpanel{(75,-5) -- (95,-5) -- (105,5) -- (125,5)} &
\payoffpanel{(75,-5) -- (95,-5) -- (105,5) -- (125,5)} \\[1.5em]

\payofftitle{\textbf{[dg]} Defined risk reversal\\
$+C_{105}-P_{95}+P_{90}$\\ zero net premium} &
\payofftitle{\textbf{[ls]} Long straddle\\
$+P_{100}+C_{100}$; debit $8$} &
\payofftitle{\textbf{[lg]} Long strangle\\
$+P_{95}+C_{105}$; debit $4$} \\[0.2em]

\payoffpanel{(75,-5) -- (90,-5) -- (95,0) -- (105,0) -- (125,20)} &
\payoffpanel{(75,17) -- (100,-8) -- (125,17)} &
\payoffpanel{(75,16) -- (95,-4) -- (105,-4) -- (125,16)} \\[1.5em]

\payofftitle{\textbf{[ib]} Iron butterfly\\
$+P_{90}-P_{100}-C_{100}+C_{110}$\\ credit $7$} &
\payofftitle{\textbf{[ic]} Iron condor\\
$+P_{90}-P_{95}-C_{105}+C_{110}$\\ credit $2$} &
\payofftitle{\textbf{[bf]} Butterfly\\
$+C_{90}-2C_{100}+C_{110}$\\ debit $3$} \\[0.2em]

\payoffpanel{(75,-3) -- (90,-3) -- (100,7) -- (110,-3) -- (125,-3)} &
\payoffpanel{(75,-3) -- (90,-3) -- (95,2) -- (105,2) -- (110,-3) -- (125,-3)} &
\payoffpanel{(75,-3) -- (90,-3) -- (100,7) -- (110,-3) -- (125,-3)} \\

\end{tabular}

\vspace{0.9em}
{\footnotesize
\begin{minipage}{\textwidth}
\textsuperscript{\,a}\,%
$C_K$ and $P_K$ denote a call and a put with strike $K$ and a common expiration
$T$; the subscript denotes the strike. Their terminal payoffs are
$(S_T-K)^+$ and $(K-S_T)^+$, where $z^+=\max(z,0)$. A plus sign denotes a
long position and a minus sign a short position; coefficients indicate contract
ratios. A \emph{debit} is the net premium paid at entry and is subtracted from
terminal payoff, while a \emph{credit} is the net premium received and is added.
All values are per share, with illustrative strikes and premiums. Vertical and
butterfly families admit the corresponding call/put and directional variants;
one representative payoff is shown for each family.
\end{minipage}}
\end{table}

% \begin{table}[tbp]
% \centering
% \footnotesize
% \caption{Leg structure and mechanism of the nine registered families}
% \label{tab:family-legs}
% \vspace{0.4em}
% \setlength{\tabcolsep}{3pt}\begin{tabularx}{\linewidth}{@{}llYl@{}}
% \toprule
% Code & Family & Legs at default coordinates & Mechanism \\
% \midrule
% \code{ol} & Outright & buy one option, \(|\delta|=0.55\) & direction \\
% \code{dv} & Debit vertical & buy \(|\delta|=0.55\), sell same right \(0.10\) farther out & direction \\
% \code{cv} & Credit vertical & sell \(|\delta|=0.35\) of the right opposite the view, buy \(0.10\) farther out & direction \\
% \code{dg} & Defined risk reversal & buy \(|\delta|=0.35\) on the view side, sell \(|\delta|=0.35\) of the opposite right, buy a \(0.20\) tail wing & direction \emph{and} skew \\
% \code{ls} & Long straddle & buy the \(0.50\) call and the \(0.50\) put & volatility \\
% \code{lg} & Long strangle & buy the \(0.25\) put and the \(0.25\) call & volatility \\
% \code{ib} & Iron butterfly & sell the \(0.50\) straddle, buy \(0.25\) wings & volatility; incidental curvature \\
% \code{ic} & Iron condor & sell the \(0.30\) strangle, buy wings \(0.10\) farther out & volatility; incidental curvature \\
% \code{bf} & Butterfly & \(1\)--\(2\)--\(1\) in one right about \(|\delta|=0.45\), widths \(0.10\) & incidental curvature \\
% \bottomrule
% \end{tabularx}
% \end{table}

\section{Point-in-time data pipeline}
\label{app:pit-pipeline}

\subsection{Market features}
\label{app:market-features}

Write \(S_t\) for the underlying price at the decision, \(r_i\) for a daily close-to-close
log return, and \(\sigma^{\mathrm{iv}}(\delta, \tau)\) for the implied volatility at
delta coordinate \(\delta\) and tenor \(\tau\) calendar days.  Unless stated
otherwise volatility is annualized for comparability.

\paragraph{Return and realized volatility.}
The step return is the log return over the session,
\begin{equation}
  \mathrm{ret}_t \;=\; \log\bigl(S_t / S_{t-1}\bigr) ,
\label{eq:feat-ret}
\end{equation}
Realized volatility is the zero-mean estimator over a trailing
\(n = 21\) trading days,
\begin{equation}
  \mathrm{rv}_t \;=\; \sqrt{\frac{252}{n}\sum_{i=1}^{n} r_{t-i}^{2}} .
\label{eq:feat-rv}
\end{equation}

\paragraph{Implied volatility and its change.}
The implied volatility is measured using the at-the-money call (with $\Delta$=0.5) with 30 days to expiry. The feature dv is the intraday change of the iv from market open to market close.
\begin{equation}
  \mathrm{iv}_t \;=\; \sigma^{\mathrm{iv}}_t(0.50\,\mathrm{C},\, 30) ,
  \qquad
  \mathrm{dv}_t \;=\; \mathrm{iv}_t - \mathrm{iv}_{t}^{\mathrm{open}} .
\label{eq:feat-iv}
\end{equation}

\paragraph{The iv wedge, term structure, skewness, and curvature.}
\begin{align}
  \mathrm{w}_t  &\;=\; \mathrm{iv}_t - \mathrm{rv}_t , \label{eq:feat-w}\\
  \mathrm{ts}_t &\;=\; \sigma^{\mathrm{iv}}_t(0.50\,\mathrm{C}, 90)
                   \;-\; \sigma^{\mathrm{iv}}_t(0.50\,\mathrm{C}, 30) ,
                   \label{eq:feat-ts}\\
  \mathrm{sk}_t &\;=\; \sigma^{\mathrm{iv}}_t(0.25\,\mathrm{P}, 30)
                   \;-\; \sigma^{\mathrm{iv}}_t(0.25\,\mathrm{C}, 30) ,
                   \label{eq:feat-sk}\\
  \mathrm{bf}_t &\;=\; \tfrac12\bigl[\sigma^{\mathrm{iv}}_t(0.25\,\mathrm{P}, 30)
                   + \sigma^{\mathrm{iv}}_t(0.25\,\mathrm{C}, 30)\bigr]
                   \;-\; \mathrm{iv}_t .
                   \label{eq:feat-bf}
\end{align}
The wedge $\mathrm{w}_t$ measures the difference between implied and realized volatility. The term structure $\mathrm{ts}_t$ measures the slope of the implied-volatility term structure as the difference between 90-day and 30-day at-the-money call implied volatility. The skewness measure $\mathrm{sk}_t$ captures the asymmetry of the implied-volatility surface as the difference between 30-day 25-delta put and call implied volatility. Finally, the curvature measure $\mathrm{bf}_t$ is the 30-day 25-delta butterfly, defined as the average implied volatility of the 25-delta put and call relative to at-the-money implied volatility. It captures the curvature of the implied-volatility smile.

\paragraph{Flow.}
The flow imbalance is the delta-weighted signed demand over the session, divided
by the total delta exposure transacted,
\begin{equation}
  \mathrm{fi}_t \;=\;
    \frac{\sum_{k} q_k \,\mathrm{sign}_k\, |\Delta_k|}{\sum_{k} q_k\,|\Delta_k|} ,
  \qquad
  \mathrm{doi}_t \;=\; \frac{\mathrm{OI}_t - \mathrm{OI}_{t-1}}{\mathrm{OI}_{t-1}} ,
\label{eq:feat-flow}
\end{equation}
the sums running over option trades \(k\) printed between the session open and the
decision time, with \(\mathrm{sign}_k \in \{+1,-1\}\) the inferred trade
direction. Trade direction is inferred from the prevailing quotes: trades above (below) the
midpoint are classified as buyer- (seller-) initiated, with midpoint trades assigned
using the most recent price change \citep{lee1991direction}.

\section{Trading and execution assumptions}
\label{app:execution}

\begin{table}[H]
\centering
\small
\caption{Trading and execution assumptions.}
\begin{tabular}{@{}ll@{}}
\toprule
Assumption & Value \\
\midrule
Traded price & Midpoint of the best bid and best ask \\
Option fee & \$0.65/contract/leg \\
Assignment fee & \$5/event \\
Share commission & \$0.005/share \\
Stock borrow & 0.5\% annually \\
Stop-loss & 60\% of entry premium \\
Ruin floor & 20\% of initial NAV \\
\bottomrule
\end{tabular}
\label{tab:execution}
\end{table}

\section{Portfolio Implementation}
\label{app:formulas}

This appendix states the three resolvers to determine contract, size,
and hedge.

\subsection{Contract resolver}
\label{app:contract-resolver}

The contract resolver maps a strategy into an exact set of contracts.  The expiry is chosen by the closest available expiry to the anchor of each tenor bucket \(a \in \{5, 14, 45, 120\}\) days. The strike price is selected by the contract with $\Delta$ exposure closest to the requested $\Delta^*$.

\subsection{Size resolver}
\label{app:sizing}

The packages in a portfolio are traded in integer units based on equal notional value, and equal risk. 

The risk is estimated by a second-order Taylor expansion of package value. The shocks are a
one-standard-deviation move in the underlying taken in whichever direction hurts, a
10\% move in implied volatility, and one day of carry:
\begin{equation}
R
=
\underbrace{|\Delta S|\,\hat{\sigma}}_{\text{delta}}
-
\underbrace{\frac{1}{2}\Gamma(S\hat{\sigma})^2}_{\text{gamma}}
+
\underbrace{|\nu|\,\sigma\,\kappa}_{\text{vega}}
-
\underbrace{\Theta}_{\text{carry}},
\qquad
\hat{\sigma}
=
\frac{\sigma^{\mathrm{atm}}}{\sqrt{252}}.
\label{eq:size-risk}
\end{equation}
with $\kappa$ the relative volatility shock.

\subsection{Hedge resolver}
\label{app:hedge-band}

Hedging is applied to non-directional strategies to offset the $\Delta$ exposure of the portfolio. To reduce transaction cost, there is a no-hedge band for small $\Delta$, 
which is given by the formula that \citet{whalley1997asymptotic} derive as
\begin{equation}
  H \;=\;
  \left(\frac{3}{2}\,
    \frac{k\,S\,\Gamma^{2}\,e^{-r(T-t)}}{\lambda}\right)^{1/3},
\label{eq:ww-band}
\end{equation}
where the hedging neutralizes the aggregated option positions' delta exposure.

\subsection{Evaluation Metrics}
\label{app:evaluation-metrics}

Let $N_t$ denote portfolio net asset value at the close of date $t$, $r_t$ the daily portfolio return, $\bar r$ its sample mean, $m=252$ the annualization factor, and $r_f$ the annualized risk-free rate.

\paragraph{Total return (TR).}
TR measures the cumulative portfolio return over the evaluation period:
$\mathrm{TR}=N_T/N_0-1$.

\paragraph{Annualized volatility (AVOL).}
AVOL measures the annualized standard deviation of daily portfolio returns:
$\mathrm{AVOL}=\sqrt{m}\sqrt{\frac{1}{T-1}\sum_{t=1}^{T}(r_t-\bar r)^2}$.

\paragraph{Maximum drawdown (MDD).}
MDD measures the largest peak-to-trough decline in portfolio net asset value:
$\mathrm{MDD}=\max_{0\leq u\leq t\leq T}(N_u-N_t)/N_u$.

\paragraph{Annualized Sharpe ratio (ASR).}
ASR measures annualized excess return relative to total volatility:
$\mathrm{ASR}=(m\bar r-r_f)/\mathrm{AVOL}$.

\paragraph{Annualized Calmar ratio (ACR).}
ACR measures annualized excess return relative to maximum drawdown:
$\mathrm{ACR}=(m\bar r-r_f)/\mathrm{MDD}$.

\paragraph{Annualized Sortino ratio (ASoR).}
ASoR measures annualized excess return relative to annualized downside deviation:
$\mathrm{ASoR}=(m\bar r-r_f)/\mathrm{DD}$,
where
$\mathrm{DD}=\sqrt{m}\sqrt{\frac{1}{T}\sum_{t=1}^{T}\min(r_t,0)^2}$.

\paragraph{Win rate (WR).}
WR is the fraction of completed trades with positive realized profit:
$\mathrm{WR}=N_{\mathrm{win}}/N_{\mathrm{total}}$.

\paragraph{Profit/loss ratio (PLR).}
PLR is the average profit of winning trades relative to the average absolute loss of losing trades:
$\mathrm{PLR}=\left|\mathbb{E}[r_i\mid r_i>0]\right|/
\left|\mathbb{E}[r_i\mid r_i<0]\right|$.

\subsection{Frontier Trajectories Performance by Strategy Family}
\label{app:family-breakdown}

\begin{table}[t]
\centering
\footnotesize
\setlength{\tabcolsep}{4.2pt}
\renewcommand{\arraystretch}{0.96}

\caption{Frontier teacher performance by strategy families.}
\label{tab:teacher-families}

\begin{tabular}{@{}llrrrrrrrr@{}}
\toprule
& & & Profit & \multicolumn{3}{c}{Risk-adjusted} & \multicolumn{2}{c}{Risk} & \\
\cmidrule(lr){4-4}
\cmidrule(lr){5-7}
\cmidrule(lr){8-9}
Code & Family & Trades
& TR(\%)
& ASR & ACR & ASoR
& AVOL(\%) & MDD(\%)
& WR(\%) \\
\midrule

\multicolumn{10}{@{}l}{\emph{Directional}} \\
\code{ol} & Outright
& 4.29 & $+3.16$ & 1.43 & 5.08 & 2.93 & 6.91 & 2.33 & 46.3 \\

\code{dv} & Debit vertical
& 2.84 & $+2.86$ & 2.10 & 8.24 & 4.60 & 5.51 & 1.68 & 75.3 \\

\code{cv} & Credit vertical
& 1.46 & $+0.73$ & 1.97 & 14.88 & 7.28 & 2.54 & 0.70 & 82.4 \\

\code{dg} & Defined risk reversal
& 0.28 & $+0.22$ & 1.31 & 4.48 & 1.99 & 4.87 & 1.51 & 60.9 \\

\addlinespace[1.5pt]
\multicolumn{10}{@{}l}{\emph{Long gamma}} \\
\code{ls} & Long straddle
& 15.00 & $-0.57$ & $-0.20$ & $-0.29$ & $-0.21$ & 20.38 & 9.75 & 30.8 \\

\code{lg} & Long strangle
& 0.69 & $+0.68$ & 0.25 & 2.46 & 1.72 & 6.20 & 1.91 & 38.3 \\

\addlinespace[1.5pt]
\multicolumn{10}{@{}l}{\emph{Short gamma}} \\
\code{ib} & Iron butterfly
& 16.23 & $+1.54$ & 0.34 & 0.73 & 0.34 & 10.44 & 4.89 & 47.4 \\

\code{ic} & Iron condor
& 5.49 & $+2.21$ & 1.47 & 4.23 & 2.44 & 7.93 & 3.20 & 62.8 \\

\code{bf} & Butterfly
& 0.02 & $-0.01$ & $-0.29$ & 3.09 & 0.20 & 1.43 & 1.02 & 33.3 \\

\midrule
\multicolumn{2}{@{}l}{Delta hedge}
& -- & $-2.83$ & $-0.82$ & $-1.06$ & $-1.16$ & 15.55 & 12.11 & -- \\

\multicolumn{2}{@{}l}{\textbf{Total}}
& -- & $\mathbf{+7.99}$ & -- & -- & -- & -- & -- & -- \\

\bottomrule
\end{tabular}

\vspace{2pt}
\begin{minipage}{\linewidth}
\footnotesize
We generate 1,000 trajectories using the frontier model and retain those with an annualized Sharpe ratio above 0.75, resulting in 144 trajectories used for SFT.
\end{minipage}

\end{table}

\clearpage
\section{Training Algorithm}
\label{app:training-algorithm}

Algorithm~\ref{alg:sota} states the two training phases: frontier-teacher strategy generation followed by supervised fine-tuning, and reinforcement learning in the portfolio environment.

\begin{algorithm}[H]
\small
\caption{SOTA Training}
\label{alg:sota}
\begin{algorithmic}[1]

\REQUIRE
Trading environment $\mathit{ENV}$;
frontier teacher $\pi_T$;
base policy $\pi_{\theta_0}$;
deterministic resolver $\mathcal{R}$;
SFT episodes $\mathcal{D}_{\mathrm{SFT}}$;
RL episodes $\mathcal{D}_{\mathrm{RL}}$;
discount factor $\gamma$

\STATE \textbf{Phase I: Frontier-Teacher Strategy Generation and SFT}

\STATE Initialize supervised trajectory set $\mathcal{D}_T \leftarrow \emptyset$

\FOR{each teacher run over the supervised-training period}

    \STATE Initialize trading environment $\mathit{ENV}$ with portfolio NAV $V_0 = \$1{,}000{,}000$

    \FOR{each episode $e \in \mathcal{D}_{\mathrm{SFT}}$, in calendar order}

        \STATE Carry the portfolio (cash, open positions, NAV) over from the previous episode, if any

        \FOR{each decision time $t$ in episode $e$}

            \STATE Observe market state $s_t$ and contemporaneous news $n_t$

            \STATE Generate frontier-teacher strategy decision
            \[
            a_t^T \sim \pi_T(\cdot \mid s_t,n_t)
            \]

            \STATE Resolve strategy decision into executable positions
            \[
            z_t^T = \mathcal{R}(a_t^T,s_t)
            \]

            \STATE Execute option and hedge trades in $\mathit{ENV}$

            \STATE Advance $\mathit{ENV}$ and observe updated portfolio state

            \STATE Store $(s_t,n_t,a_t^T)$ in $\mathcal{D}_T$

        \ENDFOR

    \ENDFOR

\ENDFOR

\STATE Supervised fine-tune the base policy on teacher-generated trajectories
\[
\theta_{\mathrm{SFT}}
=
\arg\min_{\theta}
\left[
-\sum_{(x_t,a_t^T)\in\mathcal{D}_T}
\log \pi_{\theta}(a_t^T\mid x_t)
\right]
\]

\STATE \textbf{Phase II: Reinforcement Learning}

\STATE Initialize policy $\pi_{\theta} \leftarrow \pi_{\theta_{\mathrm{SFT}}}$

\FOR{each episode $e \in \mathcal{D}_{\mathrm{RL}}$}

    \STATE Reinitialize the same trading environment $\mathit{ENV}$
    \STATE Set initial portfolio NAV $V_0 = \$1{,}000{,}000$

    \FOR{each decision time $t$ in episode $e$}

        \STATE Observe policy information $x_t$

        \STATE Generate strategy decision
        \[
        a_t \sim \pi_{\theta}(\cdot \mid x_t)
        \]

        \STATE Resolve strategy decision into executable positions
        \[
        z_t = \mathcal{R}(a_t,s_t)
        \]

        \STATE Execute option and hedge trades in $\mathit{ENV}$

        \STATE Advance $\mathit{ENV}$ and observe $V_{t+1}$ and the next state

        \STATE Compute portfolio reward
        \[
        r_t = \log V_{t+1}-\log V_t
        \]

        \STATE Store $(x_t,a_t,r_t)$ in the rollout

    \ENDFOR

    \STATE Compute discounted portfolio rewards
    \[
    G_t = \sum_{k\geq 0}\gamma^k r_{t+k}
    \]

    \STATE Update $\pi_{\theta}$ using the RL objective

\ENDFOR

\RETURN trained option-strategy policy $\pi_{\theta}$

\end{algorithmic}
\end{algorithm}

\section{Baseline and training configurations}
\label{app:hyperparameters}

This section provides the specifications for the non-LLM baselines, and hyperparameters for the LLM training. 

\subsection{GARCH rule}
We fit GARCH(1,1) on daily close-to-close log returns,
estimated by Gaussian quasi-maximum likelihood with variance targeting,
$h_{i,t} = \bar{\sigma}_i^{2}(1-\alpha-\beta) + \alpha\varepsilon_{i,t-1}^{2}
+ \beta h_{i,t-1}$. 

\begin{table}[t]
\centering
\small
\caption{GARCH rule.}
\begin{tabular}{@{}llll@{}}
\toprule
Parameter & Symbol & Value & Selection method \\
\midrule
ARCH coefficient & $\alpha$ & 0.060 & QML on estimation sample \\
GARCH coefficient & $\beta$ & 0.762 & QML on estimation sample \\
Persistence & $\alpha+\beta$ & 0.822 & implied (half-life 3.5 d) \\
Winsorisation & --- & $4\bar{\sigma}_i$ & fixed a priori \\
Forecast horizon & $H$ & 10 trading days & matches the 14-day tenor anchor \\
Estimation sample & --- & 3{,}710 obs, 2023-09-01 to 2025-02-28 & strictly pre-test \\
Long-vol entry & --- & $R \le 0.90$ & fixed; 10\% edge vs 3.2\% friction \\
Short-vol entry & --- & $R \ge 1.25$ & fixed; 25\% edge vs 10.6\% friction \\
Long-vol structure & --- & long straddle & defined-risk long gamma \\
Short-vol structure & --- & iron condor & defined-risk short gamma \\
\bottomrule
\end{tabular}
\label{tab:garch-params}
\end{table}

\subsection{Threshold rule}

We consider one fixed threshold on one feature per family, covering all nine admitted
families, and construct an equal-weighted portfolio across strategy families and underlying.

\begin{table}[t]
\centering
\small
\setlength{\tabcolsep}{4.5pt}
\caption{Threshold rule.}
\begin{tabular}{@{}llllrr@{}}
\toprule
Family & Code & Feature & Entry & Edge & Friction \\
\midrule
\multicolumn{6}{@{}l}{\emph{Participation gate, applied to every family}} \\
all & --- & $\mathrm{ts}$ & $<0$ & --- & --- \\
\addlinespace[2pt]
\multicolumn{6}{@{}l}{\emph{Volatility: $\mathrm{iv}/\mathrm{rv}$, fair value $1.0$}} \\
Long straddle & \texttt{ls} & $\mathrm{iv}/\mathrm{rv}$ & $\le 0.90$ & 10\% & 3.2\% \\
Iron butterfly & \texttt{ib} & $\mathrm{iv}/\mathrm{rv}$ & $\ge 1.20$ & 20\% & 9.1\% \\
Iron condor & \texttt{ic} & $\mathrm{iv}/\mathrm{rv}$ & $\ge 1.25$ & 25\% & 10.6\% \\
\addlinespace[2pt]
\multicolumn{6}{@{}l}{\emph{Curvature: $\mathrm{bf}/\mathrm{iv}$, fair value $0.0$}} \\
Long strangle & \texttt{lg} & $\mathrm{bf}/\mathrm{iv}$ & $\le 0$ & --- & 4.8\% \\
Butterfly & \texttt{bf} & $\mathrm{bf}/\mathrm{iv}$ & $\le 0$ & --- & n/a \\
\addlinespace[2pt]
\multicolumn{6}{@{}l}{\emph{Skew: $\mathrm{sk}/\mathrm{iv}$, fair value $0.0$}} \\
Defined-risk reversal & \texttt{dg} & $\mathrm{sk}/\mathrm{iv}$ & $\ge 0.15$ & --- & 32.9\textsuperscript{\,a} \\
Credit vertical & \texttt{cv} & $\mathrm{sk}/\mathrm{iv}$ & $\le 0$ & --- & 9.7\% \\
\addlinespace[2pt]
\multicolumn{6}{@{}l}{\emph{Momentum: $|\mathrm{ret}|$, orientation from the sign}} \\
Outright & \texttt{ol} & $|\mathrm{ret}|$ & $\ge 0.03$ & --- & 3.2\% \\
Debit vertical & \texttt{dv} & $|\mathrm{ret}|$ & $\ge 0.02$ & --- & 11.3\% \\
\bottomrule
\end{tabular}

\label{tab:threshold-params}
\end{table}
\subsection{Training hyperparameters}
\label{app:training-hparams}

\begin{table}[H]
\centering
\small
\setlength{\tabcolsep}{6pt}
\caption{Training hyperparameters for Qwen3.8-27B.}
\begin{tabular}{@{}ll@{}}
\toprule
Parameter & Qwen3.8-27B \\
\midrule

\multicolumn{2}{@{}l}{\emph{Model and supervised fine-tuning}} \\
Hugging Face revision & \code{1d4bf0f2} \\
Precision & bf16 \\
SFT episodes & $414$ \\
Maximum sequence length & $33{,}280$ \\
Train batch size & $16$ \\
Learning rate & $1\times10^{-5}$ \\
Adam $(\beta_1,\beta_2)$ & $(0.9,0.999)$ \\
Weight decay & $0.01$ \\
Epochs & $3$ \\
Optimizer steps & $75$ \\

\addlinespace[2pt]
\multicolumn{2}{@{}l}{\emph{Reinforcement learning (GRPO)}} \\
Group size & $8$ \\
Train batch size & $3$ \\
Actor learning rate & $5\times10^{-7}$ \\
KL coefficient & $0.01$ \\
PPO clip ratio & $0.2$ \\
Maximum prompt / response tokens & $8{,}192/28{,}672$ \\
Rollout temperature (train / eval) & $1.0/0$ \\
Optimizer steps & $40$ \\
Checkpoint interval & $10$ \\

\addlinespace[2pt]
\multicolumn{2}{@{}l}{\emph{Compute}} \\
Tensor parallelism & $2$ \\
GPUs $\times$ nodes & $8\times1$ \\
Seed & $0$ \\

\bottomrule
\end{tabular}

\label{tab:hyperparameters}

\vspace{2pt}
\begin{minipage}{\linewidth}
\footnotesize
\textit{Note:} The maximum sequence length of $33{,}280$ includes chat-template
control tokens in addition to the $32{,}768$-token content budget. Episodes
exceeding the content budget are excluded during corpus construction. During
RL, discounting is applied within the trajectory-level reward. Evaluation
returns are undiscounted. Checkpoints are written every $10$
optimizer steps, and each was evaluated on the test window. The results reported
throughout use the checkpoint at step $20$.
\end{minipage}

\end{table}

\subsection{Action space and portfolio construction}
\label{app:action-portfolio-params}

Table~\ref{tab:env-params} gives the parameters of the action space and of the
three deterministic resolvers.
These values are tuned on the supervised window and then held fixed for the RL training window and test window. 

\begin{table}[t]
\centering
\small
\setlength{\tabcolsep}{6pt}
\caption{Action space, sizing and hedging parameters.}
\begin{tabular}{@{}lll@{}}
\toprule
Parameter & Symbol & Value \\
\midrule
\multicolumn{3}{@{}l}{\emph{Universe and action space}} \\
Underlyings & $N$ & 10 (nine single names, one index slot) \\
Initial net asset value & $V_0$ & \$1{,}000{,}000 \\
Cash accrual & $r_f$ & 0 \\
Registered families & $|\mathcal{F}|$ & 9 \\
Tenor buckets (days to expiry) & $\mathcal{K}$ & 0--7, 8--30, 31--90, 91--180 \\
Order lines per decision & $K$ & $\le 8$ \\
Open positions, portfolio & $P_{\max}$ & 12 \\
Open positions, per underlying & $P_u$ & 3 \\
Quote-round candidates per underlying & $Q$ & 6 \\
Delta coordinate grid & $\Delta_{\rm step}$ & 0.05 \\
\addlinespace[2pt]
\multicolumn{3}{@{}l}{\emph{Size resolver}} \\
Target scenario risk, fraction of NAV & $\phi$ & 0.005 \\
Relative implied-volatility shock & $\nu$ & 0.10 \\
NAV fraction cap & $f_{\rm NAV}$ & 0.10 \\
Scenario risk cap per underlying & $\rho_u$ & 0.05 \\
Total open scenario risk cap & $\rho_{\rm tot}$ & 0.20 \\
Net dollar delta cap, fraction of NAV & $\delta_{\max}$ & 0.50 \\
Collateral buffer & $b$ & 0.05 \\
\addlinespace[2pt]
\multicolumn{3}{@{}l}{\emph{Hedge resolver}} \\
Band half-width, fraction of NAV & $\Delta_{\rm band}$ & 0.005 \\
Hedged families & --- & the five volatility families \\
Band rule; hedge target & --- & fixed; band edge \\
\addlinespace[2pt]
\multicolumn{3}{@{}l}{\emph{Marking and forced exit}} \\
Half-spread multiplier & $m_{\rm hs}$ & 1.0 (full quoted half-spread) \\
Share half-spread & $s_{\rm shr}$ & 1\,bp \\
Force-close threshold, days to expiry & $d_{\min}$ & 1 \\
Maximum tolerated stale-mark share & $\bar{s}$ & 0.25 \\
Binomial steps, American valuation & $n_{\rm tree}$ & 128 \\
Maximum mark quote age & $\tau_q$ & 6\,h \\
\bottomrule
\end{tabular}
\label{tab:env-params}
\end{table}

\subsection{Default strategy coordinates}
\label{app:coords}

A family name alone does not determine a package. Where the policy omits a delta
coordinate, the contract resolver substitutes the family default in
Table~\ref{tab:coord-params}. Any coordinate the policy does supply is
clipped to the stated range and snapped to the $\Delta_{\rm step}$ grid. The
resolver selects the contract whose delta is nearest the requested value inside
that range. The realized coordinate is returned to the policy on the receipt
line, so a request and a fill are never assumed equal.

\begin{table}[t]
\centering
\small
\setlength{\tabcolsep}{5pt}
\caption{Family default coordinates and admitted ranges.}
\begin{tabularx}{\linewidth}{@{}lYl@{}}
\toprule
Family & Default coordinates & Admitted range \\
\midrule
Outright & long delta 0.55 & $[0.20,0.80]$ \\
Debit vertical & long delta 0.55, width 0.10 & long $[0.20,0.80]$, width $[0.05,0.30]$ \\
Credit vertical & short delta 0.35, width 0.10 & short $[0.10,0.55]$, width $[0.05,0.30]$ \\
Butterfly & centre 0.45, widths 0.10/0.10 & centre $[0.25,0.70]$, widths $[0.05,0.30]$ \\
Defined-risk reversal & directional 0.35, tail wing 0.20 & directional $[0.15,0.60]$, wing $[0.05,0.35]$ \\
Long straddle & call 0.50, put 0.50 & $[0.10,0.55]$ \\
Long strangle & call 0.25, put 0.25 & $[0.10,0.55]$ \\
Iron butterfly & short delta 0.50, wing 0.25 & short $[0.10,0.55]$, wing $[0.05,0.40]$ \\
Iron condor & short delta 0.30, width 0.10 & short $[0.10,0.55]$, width $[0.05,0.30]$ \\
\bottomrule
\end{tabularx}
\begin{minipage}{\linewidth}
\footnotesize The coordinates are given by the $\Delta$ exposure of the contract.
\end{minipage}
\label{tab:coord-params}
\end{table}